\documentclass{article}

\usepackage[preprint]{neurips_2024}

\usepackage[utf8]{inputenc} 
\usepackage[T1]{fontenc}    
\usepackage{hyperref}       
\usepackage{url}            
\usepackage{booktabs}       
\usepackage{amsfonts}       
\usepackage{nicefrac}       
\usepackage{microtype}      
\usepackage{xcolor}         
\usepackage{tabularx}
\usepackage{color}
\usepackage{algorithm}
\usepackage{multirow}
\usepackage{algorithmic}
\usepackage{subcaption}
\usepackage{graphicx}
\usepackage{natbib}
\usepackage[namelimits]{amsmath}
\setcitestyle{numbers,square}

\title{Federating to Grow Transformers with Constrained Resources without Model Sharing}

\author{%
  Shikun Shen \\
  Shandong University\\
  \texttt{shikunshen@mail.sdu.edu.cn} \\
  \And
  Yifei Zou \\
  Shandong University\\
  \texttt{yfzou@sdu.edu.cn} \\
  \And
  Yuan Yuan \\
  Shandong University\\
  \texttt{yyuan@sdu.edu.cn} \\
  \And
  Yanwei Zheng\\
  Shandong University\\
  \texttt{zhengyw@sdu.edu.cn} \\
  \And
  Peng Li\\
  University of Aizu\\
  \texttt{pengli@u-aizu.ac.jp}
  \And
  Xiuzhen Cheng\\
  Shandong University\\
  \texttt{xzcheng@sdu.edu.cn} \\
  \And
  Dongxiao Yu \\
  Shandong University\\
  \texttt{dxyu@sdu.edu.cn}
}

\begin{document}
\maketitle
\begin{abstract}
  The high resource consumption of large-scale models discourages resource-constrained users from developing their customized transformers. To this end, this paper considers a federated framework named Fed-Grow for multiple participants to cooperatively scale a transformer from their pre-trained small models. Under the Fed-Grow, a Dual-LiGO (Dual Linear Growth Operator) architecture is designed to help participants expand their pre-trained small models to a transformer. In Dual-LiGO, the Local-LiGO part is used to address the heterogeneity problem caused by the various pre-trained models, and the Global-LiGO part is shared to exchange the implicit knowledge from the pre-trained models, local data, and training process of participants. Instead of model sharing, only sharing the Global-LiGO strengthens the privacy of our approach. Compared with several state-of-the-art methods in simulation, our approach has higher accuracy, better precision, and lower resource consumption on computations and communications. To the best of our knowledge, most of the previous model-scaling works are centralized, and our work is the first one that cooperatively grows a transformer from multiple pre-trained heterogeneous models with the user privacy protected in terms of local data and models. We hope that our approach can extend the transformers to the broadly distributed scenarios and encourage more resource-constrained users to enjoy the bonus taken by the large-scale transformers.
\end{abstract}
\nopagebreak
\section{Introduction}

In recent years, the transformer with billions of parameters~\cite{transformer} has shown its high performance on addressing the various complex tasks in the domains of Natural Language Processing (NLP)~\cite{bert,gpt2,gpt3} and Computer Vision (CV)~\cite{vit,swinvit,deit}.  However, the training of such large-scale transformers has a high requirement on the computing, storage, and data resources~\cite{gpt3,gpt4}, and pitilessly reject those resource-constrained users from enjoying the bonus taken by transformer~\cite{efficientTrainingSurvey}. Thus, an efficient and resource-friendly approach to obtain a transformer has been a hot topic in recent years.

\begin{figure*}[t]
  \centering
  \includegraphics[width=1\textwidth]{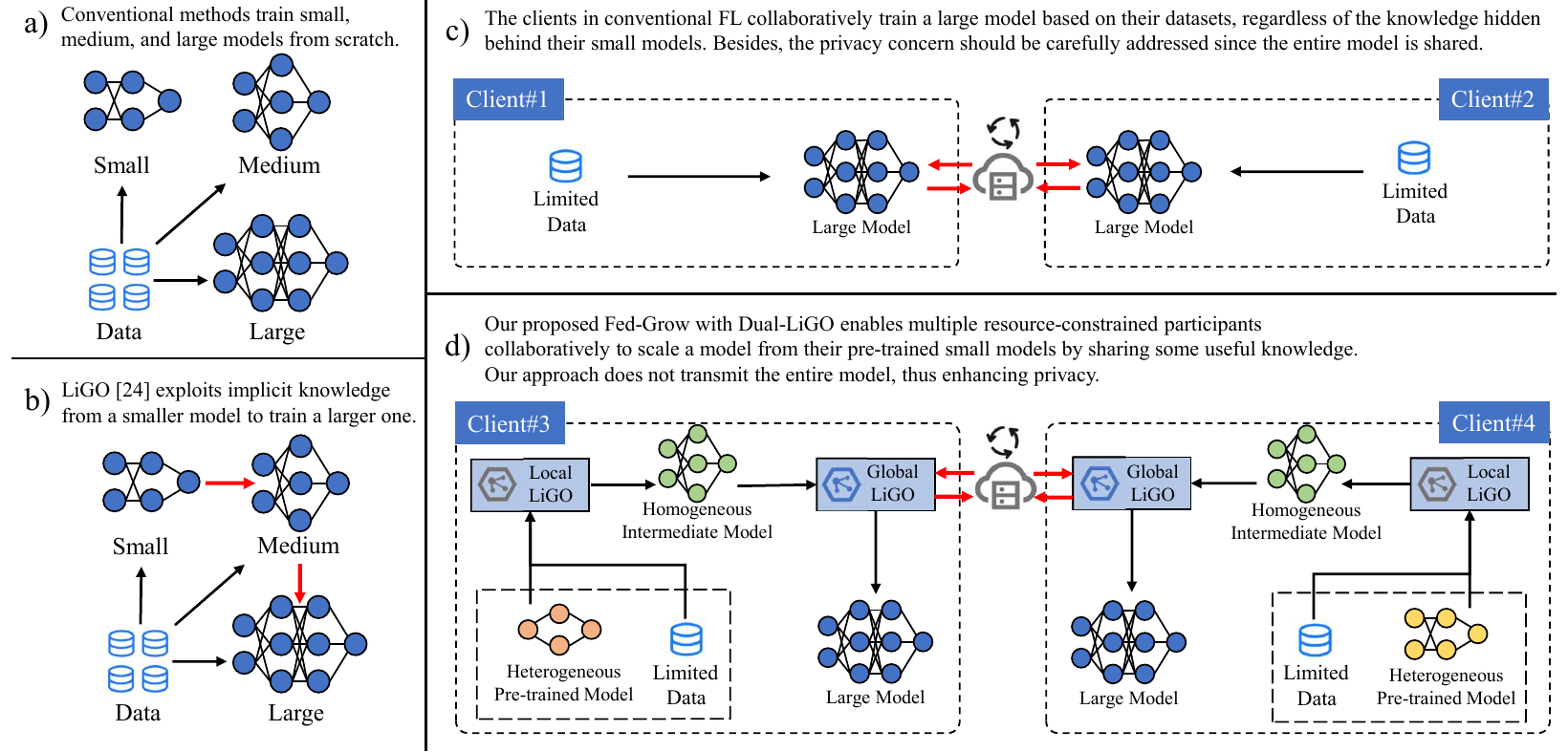}
  \caption{Illustration of different methods for efficient model training. The upper left subfigure shows the conventional methods of training a large model from scratch. The lower left subfigure shows the model reusing methods of exploiting a small pre-trained model. The upper right figure shows the conventional federated learning (FL) approach. The lower right subfigure shows our proposed framework: Fed-Grow with Dual-LiGO.}
  \label{fig:introduction}
  \vspace{-10pt}
\end{figure*}

Due to its significance, some relevant works have been proposed to reduce the resource consumption of training a transformer. Some noteworthy strategies include mixed precision training~\cite{mixedPrecision}, large batch optimization~\cite{largeBatchOptimization}, and dropping layers~\cite{layerDropping} or tokens~\cite{tokenDropping}. A common ground of the abovementioned approaches is that they all train transformers from scratch, which is not resource-saving if the targeted transformer is a scaled-up version of a pre-trained one.\footnote{For example, the GPT-2 Large model with 774 million parameters is trained from scratch despite being a scaled-up version of GPT-2 Base with 117 million parameters.} Compared with the scratch-based methods, a more efficient idea is to reuse the smaller models to initialize a larger transformer, leveraging the implicit knowledge embedded in the pre-trained models. Based on this idea, a series of works have been proposed~\cite{stackBert,mslt,bert2bert,ki,ligo,lemon,tripLe}. For instance, the Linear Growth Operator (LiGO) in~\cite{ligo} is used to learn a linear mapping from the parameters of a small model to a large one, which reduces the computational cost of scratch-based training by up to 50\%, and outperforms several model-based baselines. Figure~\ref{fig:introduction} (a) and (b) are given to illustrate the scratch-based approach and pre-trained model-based approach, respectively. Even though the existing methods have demonstrated the effectiveness of reusing smaller pre-trained models to train larger ones, all of them are centralized under the single-machine environment. The corresponding problem on the distributed side still requires further investigation.

In this paper, we study the cooperative training of a transformer with several participants, each of which has a heterogeneous pre-trained model and is resource-constrained to expand a transformer individually. For example, the data of each participant is not sufficient to train a transformer with high performance by itself. To expand a transformer from its own pre-trained model, the participants choose to share some useful knowledge with each other, as is illustrated in Figure~\ref{fig:introduction} (d). Compared with the existing scratch-based (Figure~\ref{fig:introduction} (c)) and centralized model-based approaches, our approach has several advantages. First, it allows for the integration of diverse knowledge from the pre-trained models and dataset of different participants, potentially resulting in a robust and comprehensive large model. Second, it enables distributed learning for high efficiency, which can significantly accelerate the training process by utilizing the computational resources of multiple users. Third, it offers an alternative approach for the resource-constrained users to cooperatively train a transformer.

Even though the cooperative training of transformers has such advantages, designing a distributed training algorithm is not an easy task. The biggest challenge comes from the model heterogeneity, i.e., each participant may have different architectures, sizes, and parameters of their pre-trained models. Additionally, the participants may be unwilling to share their data and even pre-trained models with others, since the models are also the property. Solving the model heterogeneity problem under the constraint of privacy protection requires a deep thinking on the following question: \textit{What should be shared in the collaborative training process for knowledge integration, distributed training acceleration, and privacy protection?}

To answer this question, we introduce  \textbf{Fed-Grow} framework with \textbf{Dual-LiGO} architecture. In Fed-Grow, multiple participants cooperatively grow a transformer from their pre-trained small models. The \textbf{Dual-LiGO} architecture enables the extension of heterogeneous small pre-trained models to large models in a federated framework and is based on the idea of LiGO (Linear Growth Operator) from~\cite{ligo} in a centralized environment. The key idea of Dual-LiGO is to design two LiGO operators: Local-LiGO and Global-LiGO. In the first, each participant uses the Local-LiGO to expand its heterogeneous pre-trained  model to a homogeneous intermediate one. Then,  the Global-LiGO is used by each participant to expand its intermediate model to a homogeneous transformer. In each local training process, gradient descent method is used to update the Local-LiGO and Global-LiGO. After multiple times local training, the Global-LiGO will be shared with others to achieve distributed training acceleration. Such a local training and federating process will be repeated for sufficient times until a transformer with high performance is obtained. Since only the Global-LiGO is shared, the local data and pre-trained model can be rarely recovered by the others, which addresses the privacy concerns in collaborative learning. Our main contributions are summarized as follows:
\begin{enumerate}
  \item We introduce Fed-Grow, a federated framework for multiple users to grow a transformer from their heterogeneous pre-trained models. To the best of our knowledge, this is the first work tackling the transformer expansion problem under a distributed and heterogeneous setting. Compared with the existing centralized works requiring sufficient data and computing resources on a single machine, Fed-Grow makes full use of the fragmented resources and the pre-trained models of multiple participants to cooperatively grow a transformer. With our Fed-Grow, those resource-constrained users who cannot afford a transformer individually can also enjoy the bonus taken by transformers.
        
  \item Under our Fed-Grow framework, the Dual Linear Growth Operator (Dual-LiGO) is designed for each participant to expand its pre-trained small model to a transformer. Specifically, the Dual-LiGO consists of two parts: Local-LiGO and Global-LiGO, both of which are trained locally but only the latter one will be shared. With the help of Global-LiGO, the participants share their knowledge in model expansion, which speeds up the training process and improves the final performance of the transformer. The Local-LiGO is designed to address the heterogeneity problem caused by pre-trained models with various architectures and sizes. Since the Local-LiGO is not shared, the adversary lacks the complete knowledge to recover the local data and pre-trained models of participants, which addresses the privacy concerns in federated learning.
        
  \item Extensive experiments are conducted to validate the effectiveness of our Fed-Grow framework with Dual-LiGO on multiple datasets. Compared with the existing  method~\cite{ligo} in stand-alone settings, Dual-LiGO boosts the accuracy or precision by average 52.43\%. Compared with the conventional federated learning that directly trains a transformer from scratch~\cite{deit,bert}, Dual-LiGO reduces the trainable parameters by 59.25\% and the communication cost by 73.01\%, achieving higher training efficiency and lower resource consumption on computation and communication.
\end{enumerate}

\textbf{Roadmap:} Section \ref{sec:relatedwork} reviews the related works.  The problem setup and our Fed-Grow framework with Dual-LiGO are presented in section \ref{sec:proposedapporach}. Section \ref{sec:experiments} reports the numerical results in simulation. In Section \ref{sec:conclusion}, we conclude this paper and discuss the future work.

\section{Related Work}\label{sec:relatedwork}

The efficient training of transformers has been a topic of interest in recent years. Initial strategies include mixed precision training~\cite{mixedPrecision}, large batch optimization~\cite{largeBatchOptimization}, progressive layer dropping~\cite{layerDropping}, token dropping~\cite{tokenDropping}, sparse training~\cite{sparseTraining}, and knowledge inheritance~\cite{ki}, which aim to lower the computational cost, memory usage, and communication overhead of training a transformer, while preserve or even improve the performance of transformer on downstream tasks. A common ground of these strategies is that they all train large models from scratch, thereby discarding the valuable knowledge acquired by their smaller counterparts. Recognizing this, recent research has also focused on reusing small pre-trained models to initialize a large model before regular training, which leverages the implicit knowledge of the small model to improve the convergence speed of the large model. For example, StackBERT~\cite{stackBert} doubles the depth of BERT model by replicating the parameters of a small model. MSLT~\cite{mslt} starts with a pre-trained small model and gradually increases the depth of the model by adding new encoder layers. Compared with the above two depth-only works,   Width-and-depth methods~\cite{net2net,bert2bert,ligo,tripLe} increase both the width and the depth of the model by adding more neurons and layers to the pre-trained small model. In detail, Net2Net~\cite{net2net} expands the width of neural networks by randomly copying neurons while preserving output values via normalization. In~\cite{bert2bert}, bert2BERT learns linear mappings to scale the pre-trained model by minimizing the task loss during model expansion. LiGO~\cite{ligo} learns to grow pre-trained transformers by factorizing the linear transformation as a composition of width- and depth-growth operators. TripLe~\cite{tripLe} partially scales a model before training, while growing the rest of the new parameters during training by copying both the warmed-up weights with the optimizer states from existing weights. By leveraging the implicit knowledge of the pre-trained small models, these model-reusing methods improve the training efficiency of large models, saving computational cost and resources.

Despite these advancements, a significant research gap remains. Most of these approaches are centralized and require considerable data and computational resources on a single machine. For example, 64 dragonfish TPUs and ImageNet-1k are required in TripLe~\cite{tripLe}, which prevents those resource-constrained users from discovering the delights of large models. Furthermore, there is a lack of relevant research in distributed scenarios. So, will it be possible for multiple resource-constrained users to cooperatively expand a transformer from their pre-trained small models is still an open question in the existing research.

\section{Fed-Grow with Dual-LiGO}\label{sec:proposedapporach}
In this section, we provide the problem setup and an overview of our approach Fed-Grow with Dual-LiGO, in which multiple resource-constrained participants collaboratively scale a transformer from their pre-trained small models. During this process, the data and pre-trained models of the participants are not shared, which enhances the privacy but also makes the design harder.

\subsection{Problem Setup}

We consider a setup with $n$ participants (also termed as clients in federated learning), and a parameter server to communicate with the clients for knowledge sharing.
Importantly, the server has no access to private data or pre-trained small models of the clients, thereby ensuring privacy in federated learning. Each client $i$ has its own data $\mathcal{D}_i=\{ x_i \in \mathbb{R}^d, y_i \in \mathbb{R} \}_{i=1}^{\vert \mathcal{D}_i\vert}$ and a pre-trained small model $\Theta_i^{(small)}$. Considering that the participants in Fed-Grow may come from different institutions, we assume that the data on each client is not independently and identically distributed (non-IID), to reflect the realistic and diverse data distributions across different clients. Besides, the pre-trained small models of clients are heterogeneous, which means the number of layers and dimensions of those small models are various. Following the notations of~\cite{ligo}, we denote the parameters of a neural network with $L$ layers and $D$ dimensions as $\Theta_{L,D}\in \mathbb{R}^{LD\times D}$. The LiGO in the original work~\cite{ligo} learns a linear mapping $\mathbf{M}: \mathbb{R}^{L_1 D_1\times D_1} \to \mathbb{R}^{L_2 D_2 \times D_2}$ from the parameters of a smaller pre-trained model $\Theta_{L_1,D_1}$ to initialize a larger one $\Theta_{L_2,D_2}$, i.e., $\Theta_{L_2,D_2}=\mathbf{M}(\Theta_{L_1,D_1}$) where $L_1 < L_2$ and $D_1 < D_2$. LiGO effectively reuses the knowledge from the smaller model to accelerate the training of the larger model.

  The goal of clients in Fed-Grow is to utilize their local model $\Theta_i^{(small)}$ to collaboratively scale a transformer $\Theta^{(large)}$ by sharing useful experience or knowledge with others. Let $\ell(\Theta;x,y)$ denote the loss function for given model $\Theta$ and data point $(x,y)$, such as the cross-entropy loss and the mean-squared loss. The underlying optimization goal of our Fed-Grow can be formalized as follows:
  \begin{equation}
    \underset{\Theta^{(large)}}{\min} \left\{ F(\Theta^{(large)}) := \sum_{i=1}^n \frac{\vert \mathcal{D}_i \vert}{\sum_{j=1}^n \vert \mathcal{D}_j \vert} F_i(\Theta^{(large)}) \right\}
  \end{equation}
  where $F_i(\Theta^{(large)})=\mathbb{E}_{(x,y)\sim \mathcal{D}_i}[\ell(\Theta^{(large)};x,y)]$ denotes the average loss of $\Theta^{(large)}$ over local dataset $\mathcal{D}_i$. The objective of the optimization problem is to find $\Theta^{(large)}$ that minimizes the weighted sum of the average losses overall local datasets, which ensures that the large model performs well on all clients' data, taking into account the size of each client's dataset.

  The important notations in problem setup and our method are presented in Appendix~\ref{appendix:notations}.


  \subsection{Overview of Our Method}
  
  Building upon the problem setup, we now turn our attention to the challenges posed by the heterogeneous model settings in a multi-user collaborative scenario. Specifically, we need to figure out what to share in the federated learning framework, to guarantee efficiency and privacy simultaneously.

  To address these challenges, we introduce the Dual-LiGO architecture. The key idea behind Dual-LiGO is to leverage two operators, the Local-LiGO and the Global-LiGO, to scale a homogeneous global model from multiple heterogeneous pre-trained models.

  In the first step, the Local-LiGO is designed to handle the heterogeneity of the small models. It expands each client's small model into a homogeneous intermediate one. This process ensures that all intermediate models have the same architecture, which is crucial for knowledge sharing in the next step.

  In the second step, the Global-LiGO takes the homogeneous intermediate models and expands them into a homogeneous large model. This operator is learned collaboratively by all clients, ensuring that the knowledge from all small models is effectively integrated into the large model without compromising the privacy of the pre-trained small model stored in each client.

  Through this two-step process, Dual-LiGO can effectively aggregate heterogeneous small models into a unified large model, while preserving the privacy of each client with respect to its small model and local data. Furthermore, by reusing the knowledge from the small models, Dual-LiGO accelerates the training process and enhances the final performance of the large model.


  \subsection{Dual-LiGO Architecture}

  The Dual-LiGO architecture consists of two main components: the Local-LiGO and the Global-LiGO.

  \textbf{Local-LiGO.} The Local-LiGO tackles the heterogeneity of the small models. For each client $i$, it expands the pre-trained small model $\Theta_i^{(small)}$ into a homogeneous intermediate model $\Theta_i^{(inter)}$ that has the same architecture across all clients. Let's denote the intermediate model as $\Theta_{L_{inter},D_{inter}} \in \mathbb{R}^{L_{inter}D_{inter}\times D_{inter}}$. The Local-LiGO in each client learns a linear mapping $\mathbf{M}_i^{(local)}:\mathbb{R}^{L_i D_i \times D_i} \to \mathbb{R}^{L_{inter}D_{inter}\times D_{inter}}$ from the pre-trained small model $\Theta_i^{(small)}$ to initialize the intermediate model $\Theta_i^{(inter)}$, i.e.,
  \begin{equation}
    \begin{aligned}
      \Theta_i^{(inter)}=\mathbf{M}_i^{(local)}(\Theta_i^{(small)}).
    \end{aligned}
  \end{equation}
  The optimization goal of the Local-LiGO is to minimize the loss on the local dataset under the constraint that the parameters of the small model are fixed. This can be formalized as follows:
  \begin{equation}
    \label{eq:local_opt_goal}
    \underset{\mathbf{M}_i^{(local)}}{\min} J(\mathbf{M}_i^{(local)}) := \left\{ \mathbb{E}_{(x,y)\sim \mathcal{D}_i}[\ell(\mathbf{M}_i^{(local)}(\Theta_i^{(small)});x,y)]\right\}
  \end{equation}
  To achieve this goal, each client updates the $\mathbf{M}_i^{(local)}$ using the following gradient descent rule:
  \begin{equation}
    \mathbf{M}_i^{(local)} \gets \mathbf{M}_i^{(local)} - \eta_l \nabla_{\mathbf{M}_i^{(local)}} J(\mathbf{M}_i^{(local)})
    \label{eq:SGD1}
  \end{equation}
  where $\eta_l$ is the learning rate for updating $\mathbf{M}_i^{(local)}$.

  \textbf{Global-LiGO.} The Global-LiGO scales the homogeneous intermediate models to a homogeneous large model. It learns a linear mapping $\mathbf{M}^{global}: \mathbb{R}^{L_{inter} D_{inter}\times D_{inter}} \to \mathbb{R}^{L_{large} D_{large} \times D_{large}}$ from the parameters of the intermediate model $\Theta_{L_{inter},D_{inter}}$ to initialize the large model $\Theta_i^{(large)}$, i.e.,
  \begin{equation}
    \begin{aligned}
      \Theta_i^{(large)}=\mathbf{M}_i^{(global)}(\Theta_i^{(inter)}).
    \end{aligned}
  \end{equation}
  The Global-LiGO is what we actually share in each communication round. By sharing this operator instead of the local models themselves, we can ensure the privacy of each client with respect to its pre-trained model and local data. Similar to Local-LiGO, the optimization goal of Global-LiGO for each client can be formulated as follows:
  \begin{equation}
    \label{eq:global_opt_goal}
    \underset{\mathbf{M}_i^{(global)}}{\min} J(\mathbf{M}_i^{(global)}) := \left\{ \mathbb{E}_{(x,y)\sim \mathcal{D}_i}[\ell(\mathbf{M}_i^{(global)}(\Theta_i^{(inter)});x,y)]\right\}
  \end{equation}
  Likewise, each client updates the $\mathbf{M}_i^{(global)}$ using the following gradient descent rule:
  \begin{equation}
    \mathbf{M}_i^{(global)} \gets \mathbf{M}_i^{(global)} - \eta_g \nabla_{\mathbf{M}_i^{(global)}} J(\mathbf{M}_i^{(global)})
    \label{eq:SGD2}
  \end{equation}
  where $\eta_g$ is the learning rate for updating $\mathbf{M}_g^{(global)}$.

  The workflow of Dual-LiGO is illustrated in Figure~\ref{fig:method}, and its pseudocode is given in Appendix~\ref{appendix:pseudocode}. For each client, it firstly train the $\mathbf{M}_i^{(local)}$ locally for $E_l$ epochs, and get the intermediate homogeneous model via $\mathbf{M}_i^{(local)}$. Next, it trains Global-LiGO locally for $E_g$ epochs and sends the Global-LiGO to the parameter server for aggregation. The aggregated Global-LiGO will be used as the starting point of the next $E_g$ epochs local training. Such a local training and aggregation process will be repeated for $R$ times. Finally, based on the intermediate model and Global-LiGO, each client gets a homogeneous transformer.

  \begin{figure*}[ht]
    \centering
    \includegraphics[width=1\textwidth]{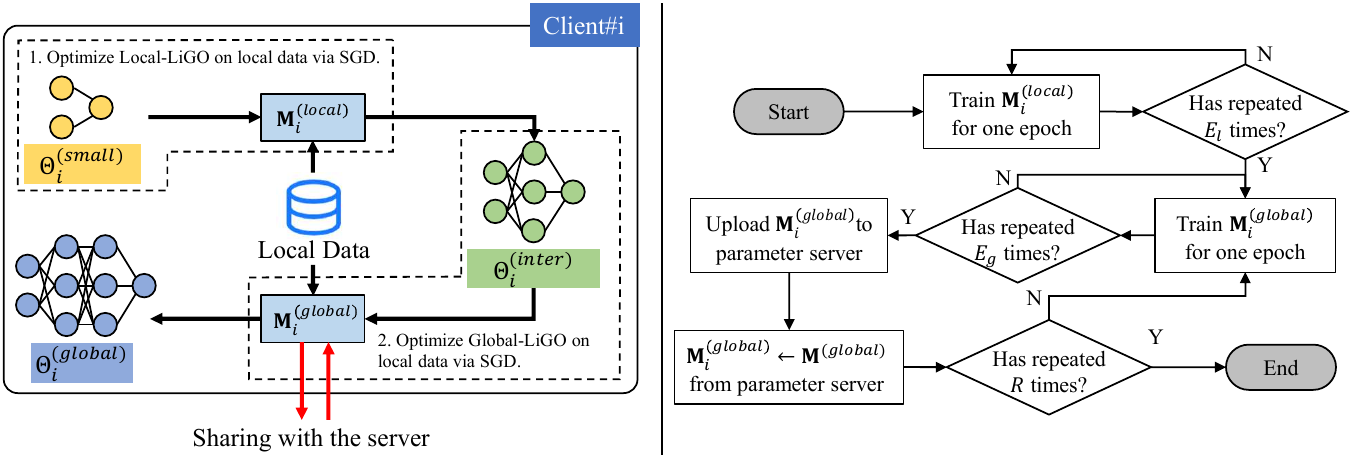}
    \caption{Workflow of Dual-LiGO.}
    \label{fig:method}
    \vspace{-10pt}
  \end{figure*}

  \section{Experiments}\label{sec:experiments}

  In this section, we first introduce the experiment setup and then show the numerical results demonstrating the effectiveness of our Fed-Grow with Dual-LiGO.

  \subsection{Datasets and Models}

  We consider the following 3 tasks: text classification, sequence tagging and image classification. For text classification tasks, we use the 20 Newsgroups~\cite{20news} and AG News datasets~\cite{agnews}, employing the BERT~\cite{bert} model. For sequence tagging tasks, we use the WikiNER dataset~\cite{wikiner}, also utilizing the BERT~\cite{bert} model. For image classification tasks, we use the CIFAR-10~\cite{cifar10and100}, CIFAR-100~\cite{cifar10and100}, and Flowers-102~\cite{flowers102} datasets, using the Vision Transformer (ViT)~\cite{vit} model. The detailed description about the dataset is available in the Appendix~\ref{appendix:datasetintro}

  For each client's small model, i.e., $\Theta^{(small)}$, we design three different sizes to simulate the heterogeneous setting of the small models. Details about the sizes of small models $\Theta^{(small)}$, intermediate models $\Theta^{(inter)}$, and large models $\Theta^{(large)}$ are provided in Appendix~\ref{appendix:modelconfig}.
  \begin{table*}[!ht]
    \caption{Comparison of accuracy/precision and standard deviation between Fed-Grow (Agg) and NoAgg~\cite{ligo} under different settings.}
    
    \resizebox{1.0\textwidth}{!}{%
      \begin{tabular}{cccccccccccccc}
        \toprule
                                &                        & \multicolumn{12}{c}{\textbf{Accuracy/Precision (\%)} $\uparrow$}                                                                                                                                                                                                                                                \\ \cmidrule{3-14}
                                &                        & \multicolumn{6}{c}{10 Clients}                                   & \multicolumn{6}{c}{20 Clients}                                                                                                                                                                                                               \\ \cmidrule{3-14}
                                &                        & \multicolumn{3}{c}{IID}                                          & \multicolumn{3}{c}{Non-IID}    & \multicolumn{3}{c}{IID}        & \multicolumn{3}{c}{Non-IID}                                                                                                                                                \\ \cmidrule{3-14}
        \multirow{-5}{*}{Model} & \multirow{-5}{*}{Task} & Agg                                                              & NoAgg                          & $\Delta$                       & Agg                         & NoAgg  & $\Delta$                       & Agg    & NoAgg  & $\Delta$                       & Agg    & NoAgg  & $\Delta$                      \\ \midrule
                                & 20news                 & 65.295                                                           & 49.177                         & {\color[HTML]{FF0000} 16.118}  & 53.983                      & 37.759 & {\color[HTML]{FF0000} 16.224}  & 48.258 & 45.470 & {\color[HTML]{FF0000} 2.788}   & 27.787 & 18.693 & {\color[HTML]{FF0000} 9.094}  \\ \cmidrule{2-14}
                                & agnews                 & 90.158                                                           & 87.474                         & {\color[HTML]{FF0000} 2.684}   & 85.671                      & 73.079 & {\color[HTML]{FF0000} 12.592}  & 86.053 & 86.737 & {\color[HTML]{FF0000} -0.684}  & 84.250 & 49.303 & {\color[HTML]{FF0000} 34.947} \\ \cmidrule{2-14}
        \multirow{-4}{*}{BERT}  & wikiner                & 42.581                                                           & 28.324                         & {\color[HTML]{FF0000} 14.257}  & 43.878                      & 33.539 & {\color[HTML]{FF0000} 10.339}  & 43.693 & 28.803 & {\color[HTML]{FF0000} 14.890}  & 42.558 & 27.300 & {\color[HTML]{FF0000} 15.258} \\ \midrule
                                & CIFAR-10               & 76.717                                                           & 67.977                         & {\color[HTML]{FF0000} 8.740}   & 70.186                      & 48.771 & {\color[HTML]{FF0000} 21.415}  & 78.223 & 70.020 & {\color[HTML]{FF0000} 8.203}   & 71.914 & 53.730 & {\color[HTML]{FF0000} 18.184} \\ \cmidrule{2-14}
                                & CIFAR-100              & 53.178                                                           & 38.127                         & {\color[HTML]{FF0000} 15.051}  & 50.109                      & 26.904 & {\color[HTML]{FF0000} 23.205}  & 54.004 & 39.141 & {\color[HTML]{FF0000} 14.863}  & 51.270 & 28.457 & {\color[HTML]{FF0000} 22.813} \\ \cmidrule{2-14}
        \multirow{-4}{*}{ViT}   & Flowers102             & 38.555                                                           & 25.000                         & {\color[HTML]{FF0000} 13.555}  & 37.928                      & 20.326 & {\color[HTML]{FF0000} 17.602}  & 31.548 & 12.448 & {\color[HTML]{FF0000} 19.100}  & 31.904 & 10.349 & {\color[HTML]{FF0000} 21.555} \\ \midrule
                                &                        & \multicolumn{12}{c}{\textbf{Standard Deviation} $\downarrow$}                                                                                                                                                                                                                                                   \\ \cmidrule{3-14}
                                &                        & \multicolumn{6}{c}{10 Clients}                                   & \multicolumn{6}{c}{20 Clients}                                                                                                                                                                                                               \\ \cmidrule{3-14}
                                &                        & \multicolumn{3}{c}{IID}                                          & \multicolumn{3}{c}{Non-IID}    & \multicolumn{3}{c}{IID}        & \multicolumn{3}{c}{Non-IID}                                                                                                                                                \\ \cmidrule{3-14}
        \multirow{-5}{*}{Model} & \multirow{-5}{*}{Task} & Agg                                                              & NoAgg                          & $\Delta$                       & Agg                         & NoAgg  & $\Delta$                       & Agg    & NoAgg  & $\Delta$                       & Agg    & NoAgg  & $\Delta$                      \\ \midrule
                                & 20news                 & 3.546                                                            & 24.219                         & {\color[HTML]{FF0000} -20.673} & 11.960                      & 23.161 & {\color[HTML]{FF0000} -11.201} & 3.518  & 14.605 & {\color[HTML]{FF0000} -11.087} & 3.518  & 6.112  & {\color[HTML]{FF0000} -2.594} \\ \cmidrule{2-14}
                                & agnews                 & 1.001                                                            & 4.362                          & {\color[HTML]{FF0000} -3.361}  & 4.350                       & 28.025 & {\color[HTML]{FF0000} -23.675} & 2.100  & 1.573  & {\color[HTML]{00B050} 0.527}   & 14.605 & 16.340 & {\color[HTML]{FF0000} -1.735} \\ \cmidrule{2-14}
        \multirow{-4}{*}{BERT}  & wikiner                & 3.513                                                            & 9.069                          & {\color[HTML]{FF0000} -5.556}  & 1.360                       & 11.288 & {\color[HTML]{FF0000} -9.928}  & 2.441  & 9.689  & {\color[HTML]{FF0000} -7.248}  & 2.241  & 11.109 & {\color[HTML]{FF0000} -8.868} \\ \midrule
                                & CIFAR-10               & 1.482                                                            & 2.908                          & {\color[HTML]{FF0000} -1.426}  & 1.870                       & 5.168  & {\color[HTML]{FF0000} -3.298}  & 1.251  & 2.280  & {\color[HTML]{FF0000} -1.029}  & 1.321  & 5.225  & {\color[HTML]{FF0000} -3.904} \\ \cmidrule{2-14}
                                & CIFAR-100              & 1.045                                                            & 1.848                          & {\color[HTML]{FF0000} -0.803}  & 1.332                       & 3.099  & {\color[HTML]{FF0000} -1.767}  & 1.401  & 1.795  & {\color[HTML]{FF0000} -0.394}  & 1.551  & 2.282  & {\color[HTML]{FF0000} -0.731} \\ \cmidrule{2-14}
        \multirow{-4}{*}{ViT}   & Flowers102             & 1.419                                                            & 1.657                          & {\color[HTML]{FF0000} -0.238}  & 1.399                       & 2.609  & {\color[HTML]{FF0000} -1.210}  & 1.467  & 2.305  & {\color[HTML]{FF0000} -0.838}  & 1.038  & 1.638  & {\color[HTML]{FF0000} -0.600} \\ \bottomrule
      \end{tabular}}
    \label{tab:mainResults}
    
  \end{table*}

  \begin{figure*}[!ht]
    \centering
    \begin{subfigure}{0.32\linewidth}
      \includegraphics[width=\linewidth]{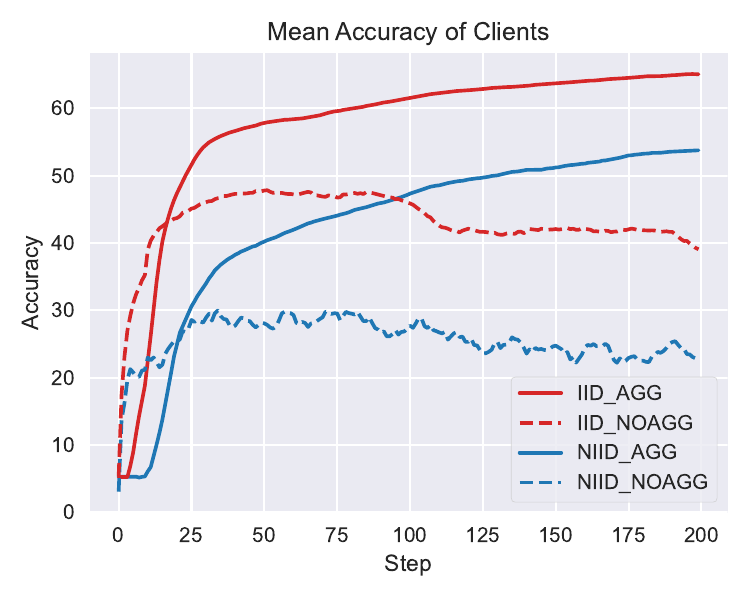}
      
      \caption{20news}
    \end{subfigure}
    \begin{subfigure}{0.32\linewidth}
      \includegraphics[width=\linewidth]{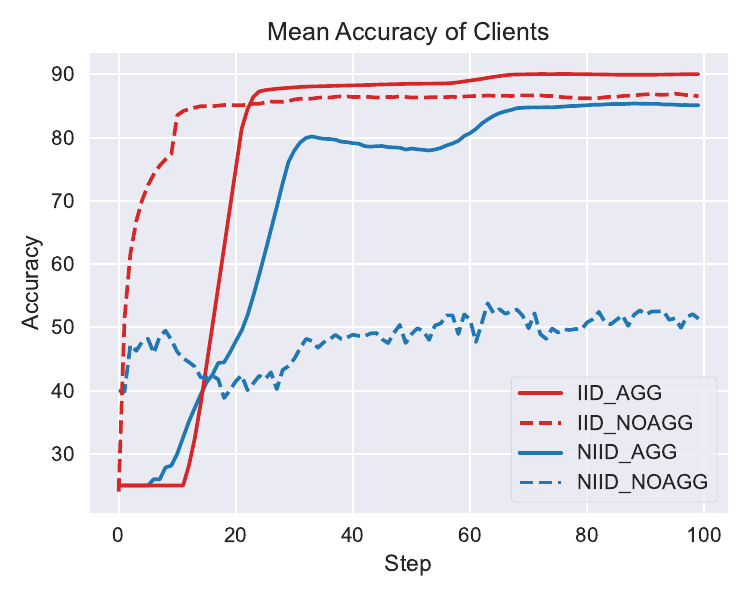}
      
      \caption{agnews}
    \end{subfigure}
    \begin{subfigure}{0.32\linewidth}
      \includegraphics[width=\linewidth]{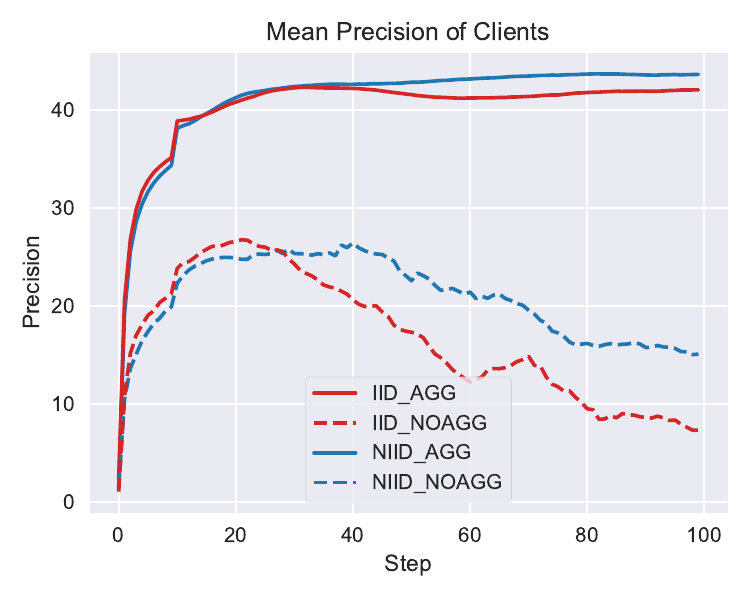}
      
      \caption{wikiner}
    \end{subfigure}
    \begin{subfigure}{0.32\linewidth}
      \includegraphics[width=\linewidth]{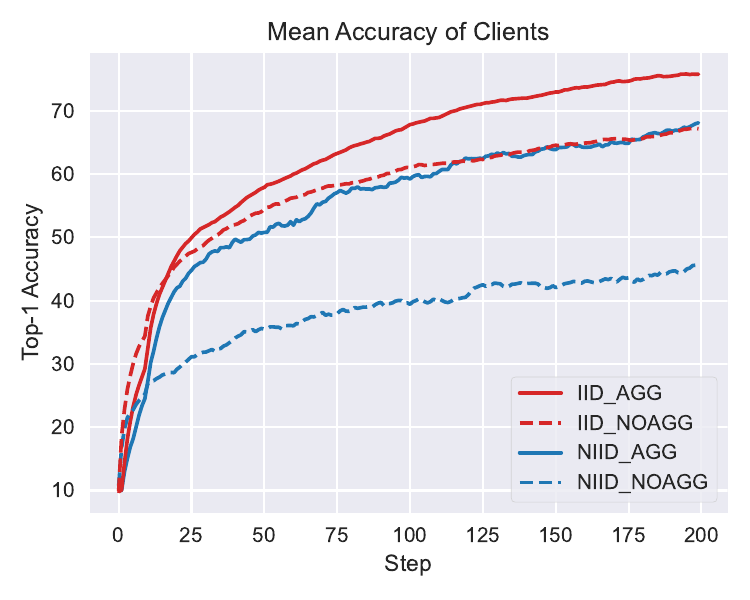}
      
      \caption{CIFAR-10}
    \end{subfigure}
    \begin{subfigure}{0.32\linewidth}
      \includegraphics[width=\linewidth]{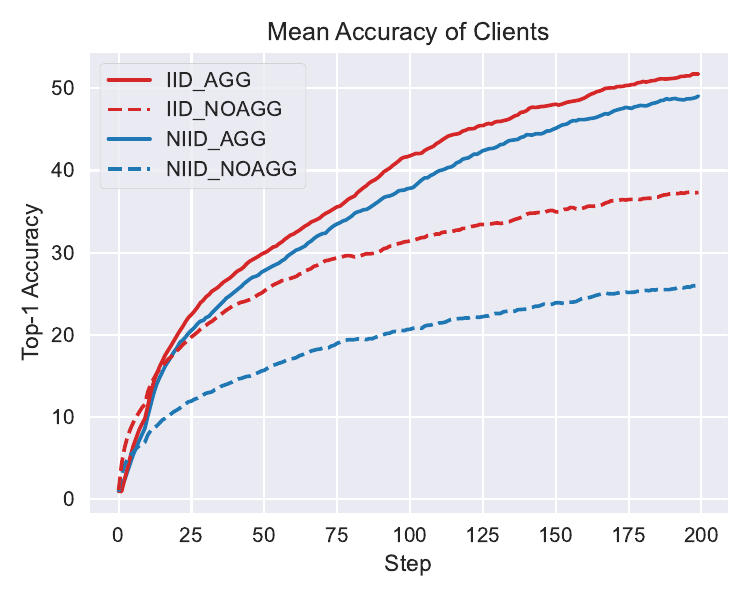}
      
      \caption{CIFAR-100}
    \end{subfigure}
    \begin{subfigure}{0.32\linewidth}
      \includegraphics[width=\linewidth]{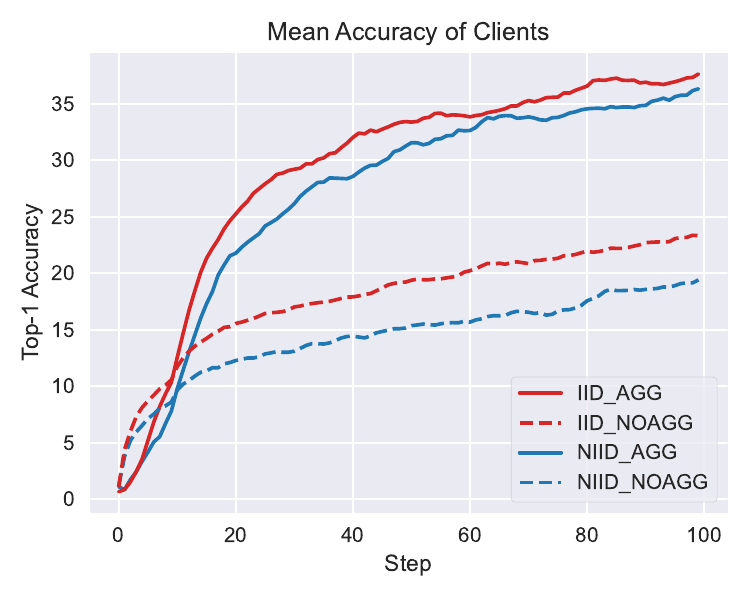}
      
      \caption{Flowers-102}
    \end{subfigure}
    \caption{Comparison of Dual-LiGO (Agg) and NoAgg on six datasets under IID and non-IID settings with 10 clients. The subfigures show the mean accuracy (or precision) of the methods on each dataset. Each subfigure shows four lines: IID\_AGG, IID\_NoAgg, NIID\_AGG, NIID\_NoAgg, which represent the test accuracy (or precision) of the methods under different settings.}
    \label{fig:mainResults}
    \vspace{-10pt}
  \end{figure*}


  \subsection{Results Overview}\label{sec:resultsAndAnalysis}

  \textbf{Baselines.} We compare our proposed method, Fed-Grow with Dual-LiGO, against two baseline methods. The first one is each client's own expansion of the large model (without aggregation)~\cite{ligo}, denoted as "NoAgg" in the tables. This baseline represents the scenario where each client uses its own pre-trained model to initialize a large model and trains it locally without any communication or knowledge sharing with other clients. We compare this baseline with our method on the top-1 accuracy or precision metrics, to show that our Fed-Grow framework really works on improving accuracy, precision, and training speed. The second one is the conventional federated learning that directly trains a large transformer from scratch~\cite{vit,bert}, denoted as "Scratch" in the tables. This baseline represents the scenario where each client starts with a randomly initialized large model and trains with federated averaging on the large model. We compare this baseline with our method on the number of trainable parameters and communication cost metrics, to show that our Dual-LiGO architecture is resource saving than the full model sharing.

  \textbf{Performance Analysis.}
  We test our models with different numbers of clients, specifically 10 and 20 clients. We analyze the performance of our Fed-Grow from three aspects: accuracy/precision, stability, and resource consumption. In Table~\ref{tab:mainResults} and Figure~\ref{fig:mainResults}, we compare our method with the NoAgg method on the accuracy and stability metrics. The results on the number of trainable parameters and communication cost metrics are shown in Table~\ref{tab:totalTrainableParamsAndCommCost}, which compares our method with the Scratch method. We also show the average test accuracy/precision over communication rounds under different settings in Appendix~\ref{appendix:learningcurves}.

  \textbf{Fed-Grow improves accuracy/precision.} Our proposed method, Fed-Grow with Dual-LiGO, generally outperforms the baseline method NoAgg~\cite{ligo} in terms of accuracy and precision across the absolute majority of datasets, both under IID and non-IID conditions, and for both 10 and 20 clients. This is indicated by the positive $\Delta$ values in the accuracy/precision part of Table~\ref{tab:mainResults}, which represent the improvement in accuracy achieved by Fed-Grow over the baseline method. There is only 1  exception to be reported. In the agnews dataset under IID conditions with 20 clients, the NoAgg method achieves a slightly higher accuracy (86.737) than Fed-Grow (86.053) and a lower standard deviation. This could be attributed to the relatively simple nature of the agnews dataset, which has only 4 classes of data. In the IID setting, every client has enough amount of data for training a robust model, which might have given the NoAgg method an advantage. The above results suggest that Fed-Grow with  Dual-LiGO can effectively integrate the diverse knowledge from different participants and achieve a robust and comprehensive large model.

  \textbf{Fed-Grow reduces the standard deviation of clients' performance.} In terms of stability, measured by the standard deviation of accuracy/precision across clients, Dual-LiGO also appears to perform better than NoAgg. This is indicated by the negative $\Delta$ values in the standard deviation part of Table~\ref{tab:mainResults}, which suggests that Fed-Grow with  Dual-LiGO has a lower standard deviation, and hence more stable performance, compared to the baseline method. This suggests that Dual-LiGO can reduce the performance gap among different clients and ensure a consistent quality of the model.

  \begin{table}[!ht]
    \vspace{-10pt}
    \centering
    \caption{Comparison on number of trainable parameters and communication cost between Dual-LiGO and Scratch~\cite{bert,vit}. The table shows the results for different components. The communication cost is measured by the number of parameters that need to be transmitted between each device and the server in each training round.}
    
    \resizebox{0.6\textwidth}{!}{\begin{tabular}{cccc}
        \toprule
        \textbf{Method}                                  & \textbf{Model}                 & \textbf{\# Trainable Params}   & \textbf{Comm Cost} \\ \midrule
                                                         & Small                          & 0                         & 0                  \\
                                                         & Local-LiGO                     & 0.737M                         & 0                  \\
        \multirow{-3}{*}{Dual-LiGO\#1}                   & Global-LiGO                    & 2.089M                         & 4.178M             \\ \midrule
                                                         & Small                          & 0                         & 0                  \\
                                                         & Local-LiGO                     & 1.065M                         & 0                  \\
        \multirow{-3}{*}{Dual-LiGO\#2}                   & Global-LiGO                    & 2.089M                         & 4.178M             \\ \midrule
                                                         & Small                          & 0                         & 0                  \\
                                                         & Local-LiGO                     & 1.393M                         & 0                  \\
        \multirow{-3}{*}{Dual-LiGO\#2}                   & Global-LiGO                    & 2.089M                         & 4.178M             \\ \midrule
        Scratch                                          & Large                          & 7.740M                         & 15.480M            \\ \midrule
        \multicolumn{2}{c}{Averaged result of Dual-LiGO} & 3.154M                         & 4.178M                                              \\ \midrule
        \multicolumn{2}{c}{Improvements on Scratch}      & {\color[HTML]{FF0000} 59.25\%} & {\color[HTML]{FF0000} 73.01\%}                      \\ \bottomrule
      \end{tabular}}
    \label{tab:totalTrainableParamsAndCommCost}
    
  \end{table}

  \textbf{Dual-LiGO reduces the resource consumption on computation and communication.} As shown in Table \ref{tab:totalTrainableParamsAndCommCost}, Dual-LiGO has fewer trainable parameters and lower communication cost than Scratch. Specifically, Fed-Grow with Dual-LiGO reduces the number of trainable parameters and the communication cost by 59.25\% and 73.01\%, respectively, which indicates that our Fed-Grow with Dual-LiGO requires less computational resources and transfers less data in the training process than Scratch. This can reduce the hardware thresholds for the participants, save bandwidth and energy cost in training, and also improve the privacy and security of the data since less parameters are shared. These results demonstrate that our Fed-Grow with Dual-LiGO is more efficient and resource-friendly than Scratch.

  In summary, our experimental results have shown that our method has three main advantages over the baseline methods. First, Fed-Grow with Dual-LiGO improves the accuracy and precision of the model across various tasks and conditions, indicating higher performance. Second, Fed-Grow with Dual-LiGO enhances the stability of the model's performance across different clients, indicating more consistency. Third, Fed-Grow with  Dual-LiGO reduces the resource consumption on computation and communication, indicating lower threshold and more green for training a transformer. These findings suggest that our Fed-Grow with Dual-LiGO is a promising approach, especially for those users with limited data and computing resources.

  \subsection{Detailed Analysis}

  In this part, we analyze the detailed performance of Fed-Grow with Dual-LiGO under different client settings. Additionally, the resource consumption of Fed-Grow with Dual-LiGO on computation and communication are also observed.

  \begin{table}[!ht]
    \vspace{-10pt}
    \centering
    \caption{Mean accuracy of Dual-LiGO (Agg) and NoAgg on the CIFAR-100 dataset under IID and non-IID settings with different numbers of clients.}

    \resizebox{0.6\textwidth}{!}{\begin{tabular}{ccccccc}
        \toprule
                                                 & \multicolumn{6}{c}{\textbf{Accuracy (\%)} $\uparrow$}                                                                                                                 \\ \cmidrule{2-7}
                                                 & \multicolumn{3}{c}{IID}                               & \multicolumn{3}{c}{Non-IID}                                                                                   \\ \cmidrule{2-7}
        \multirow{-4}{*}{\textbf{Client Number}} & Agg                                                   & NoAgg                       & $\Delta$                      & Agg    & NoAgg  & $\Delta$                      \\ \midrule
        10                                       & 53.178                                                & 38.127                      & {\color[HTML]{FF0000} 15.051} & 50.109 & 26.904 & {\color[HTML]{FF0000} 23.205} \\
        20                                       & 54.004                                                & 39.141                      & {\color[HTML]{FF0000} 14.863} & 51.270 & 28.457 & {\color[HTML]{FF0000} 22.813} \\
        30                                       & 47.57                                                 & 31.341                      & {\color[HTML]{FF0000} 16.229} & 43.727 & 21.363 & {\color[HTML]{FF0000} 22.364} \\
        40                                       & 42.061                                                & 28.499                      & {\color[HTML]{FF0000} 13.562} & 37.209 & 18.48  & {\color[HTML]{FF0000} 18.729} \\
        50                                       & 37.478                                                & 32.854                      & {\color[HTML]{FF0000} 4.624}  & 25.922 & 17.122 & {\color[HTML]{FF0000} 8.800}  \\ \bottomrule
      \end{tabular}}
    \vspace{-15pt}
    \label{tab:ablationClientNumCifar100}
  \end{table}

  \begin{table}[!ht]

    \centering
    \caption{Mean accuracy of Dual-LiGO (Agg) and NoAgg on the agnews dataset under IID and non-IID settings with different numbers of clients.}

    \resizebox{0.6\textwidth}{!}{\begin{tabular}{ccccccc}
        \toprule
                                                 & \multicolumn{6}{c}{\textbf{Accuracy (\%)} $\uparrow$}                                                                                                                 \\ \cmidrule{2-7}
                                                 & \multicolumn{3}{c}{IID}                               & \multicolumn{3}{c}{Non-IID}                                                                                   \\ \cmidrule{2-7}
        \multirow{-4}{*}{\textbf{Client Number}} & Agg                                                   & NoAgg                       & $\Delta$                      & Agg    & NoAgg  & $\Delta$                      \\ \midrule
        10                                       & 90.158                                                & 87.474                      & {\color[HTML]{FF0000} 2.684}  & 85.671 & 73.079 & {\color[HTML]{FF0000} 12.592} \\
        20                                       & 86.053                                                & 86.737                      & {\color[HTML]{00B050} -0.684} & 84.250 & 49.303 & {\color[HTML]{FF0000} 34.947} \\
        30                                       & 88.000                                                & 87.197                      & {\color[HTML]{FF0000} 0.803}  & 87.632 & 50.263 & {\color[HTML]{FF0000} 37.369} \\
        40                                       & 87.329                                                & 85.737                      & {\color[HTML]{FF0000} 1.592}  & 82.961 & 38.579 & {\color[HTML]{FF0000} 44.382} \\
        50                                       & 70.232                                                & 66.484                      & {\color[HTML]{FF0000} 3.748}  & 68.126 & 34.137 & {\color[HTML]{FF0000} 33.989} \\ \bottomrule
      \end{tabular}}
      \vspace{-10pt}
    \label{tab:ablationClientNumAgnews}
  \end{table}

  \textbf{Different Number of Clients.} We conduct a robustness analysis to test the performance of Fed-Grow with Dual-LiGO under different client settings. We use two datasets, CIFAR-100 and agnews, and test Fed-Grow  under both IID and non-IID data distributions. We vary the number of clients from 10 to 50. The results are shown in Tables \ref{tab:ablationClientNumCifar100} and \ref{tab:ablationClientNumAgnews} and the accuracy curves are provided in Appendix~\ref{appendix:learningcurves}, which show that Fed-Grow with Dual-LiGO is robust to different client settings, as it consistently outperforms baseline in both IID and Non-IID settings, except for one case where baseline has slightly higher accuracy than Fed-Grow when the number of clients is 20 in the IID setting of agnews. This has been explained in Section~\ref{sec:resultsAndAnalysis}. In most cases, Fed-Grow has a clear advantage over baseline, especially in the Non-IID setting, where the gap is very large. The largest improvement of Fed-Grow with Dual-LiGO over baseline is achieved when the number of clients is 40, with 1.592\% and 44.382\% higher accuracy in IID and Non-IID settings of agnews, respectively. The smallest improvement is observed when the number of clients is 20, with -0.684\% and 34.947\% higher accuracy in IID and Non-IID settings of agnews, respectively. This demonstrates that a client with Dual-LiGO can effectively leverage the information from other clients to improve the performance of its transformer, while the baseline suffers from limited and heterogeneous data. The numerical results shows that Dual-LiGO is robust to data heterogeneity and sparsity and can adapt to different client settings.

  \textbf{Detailed resource consumption on Dual-LiGO.}

  In our approach, each client only has to store its small model, and train the Local-LiGO and Global-LiGO to obtain its transformer. Since there are three kind of small models (\#1, \#2, \#3) in our simulation, the Dual-LiGO also vary on clients with small models \#1, \#2, \#3 with respect to trainable parameters and communication cost. The observed trainable parameters and communication cost for the Dual-LiGO\#1, Dual-LiGO\#2, Dual-LiGO\#3  are presented in the Table~\ref{tab:totalTrainableParamsAndCommCost}, from which we can see that the average trainable parameters and communication cost for Dual-LiGO are 3.154M and 4.178M, respectively. Compared with the results $7.740$M and $15.480$M for Scratch, our Dual-LiGO has $59.25\%$ and $73.01\%$ improvements on trainable parameters and communication cost.

  \section{Conclusion}\label{sec:conclusion}

  To make the resource-constrained users can also afford the costly-training transformers, this paper studies a federated framework named Fed-Grow, in which multiple participants cooperatively expand a large-scale transformer from their pre-trained small models. To efficiently exchange the useful knowledge and protect the privacy of users in model scaling, the Dual-LiGO architecture is designed based on Fed-Grow, with its Local-LiGO part used to tackle the heterogeneity problem caused by the pre-trained models, and its Global-LiGO part shared by participants to exchange the implicit knowledge from the pre-trained models and the experience learned in model scaling.
  Compared with the classical federated learning, our approach no longer shares the models, let alone the raw data, which strengthens the privacy of our approach. The numerical results from the simulation show that our approach outperforms several baselines in terms of accuracy, precision, stability, and resource consumption on computations and communications. Extending our work to train customized transformers with heterogeneous architecture will be our work in the future.

  \medskip

  \bibliographystyle{ACM-Reference-Format}
  \bibliography{ref}


  \appendix

  \section{Detailed Description of Dual-LiGO}

  \subsection{Notations}\label{appendix:notations}
  The notations to articulate the Dual-LiGO framework within our federated learning context are shown in Table~\ref{tab:notations}.

  \begin{table}[!ht]
    \centering
    \caption{Table of notations.}

    \begin{tabularx}{1\textwidth}{lX}
      \toprule
      \textbf{Notation}         & \textbf{Description}                                         \\
      \midrule
      $n$                       & Number of clients                                            \\
      $\mathcal{D}_i$           & Dataset of client $i$                                        \\
      $\Theta_{L,D}$            & A neural network with $L$ layers and $D$ dimensions          \\
      $\Theta_i^{(small)}$      & Pre-trained small model of client $i$                        \\
      $\Theta_i^{(inter)}$      & Intermediate model of client $i$                             \\
      $\Theta_i^{(large)}$      & Large model of client $i$                                    \\
      $\mathbf{M}_i^{(local)}$  & Local-LiGO of client $i$                                     \\
      $\mathbf{M}_i^{(global)}$ & Global-LiGO of client $i$                                    \\
      $\ell(\Theta;x,y)$        & Loss function given model $\Theta$ and data point $(x,y)$    \\
      $F(\Theta^{(large)})$     & Objective function of the large model                        \\
      $F_i(\Theta^{(large)})$   & Average loss of the large model over dataset $\mathcal{D}_i$ \\
      $\eta_l$                  & The learning rate of Local-LiGO                              \\
      $\eta_g$                  & The learning rate of Global-LiGO                             \\
      $E_l$                     & Number of epochs for training the Local-LiGO                 \\
      $E_g$                     & Number of epochs for training the Global-LiGO                \\
      $R$                       & Number of rounds for aggregation round                       \\
      \bottomrule
    \end{tabularx}
    \label{tab:notations}
  \end{table}

  \subsection{Pseudocode of Dual-LiGO}\label{appendix:pseudocode}
  The pseudocode for the implementation of the Dual-LiGO framework within the Fed-Grow federated learning system is shown in Algorithm~\ref{algo:workflowServer} and Algorithm~\ref{algo:workflowClient}. The algorithms detail the iterative process of training, updating, and aggregating the models, which culminates in the construction of a robust and large global model.

  \begin{algorithm}[!ht]
    \caption{Fed-Grow with Dual-LiGO for Client $i$}
    \begin{algorithmic}[1]
      \STATE \textbf{Input:}  Pre-trained small model $\Theta_i^{(small)}$;
      \STATE \qquad\quad Local dataset $\mathcal{D}_i$;  Local-LiGO training epoch $E_l$;
      \STATE \qquad\quad Federated round $R$; Global-LiGO training epoch $E_g$;
      \STATE \textbf{Output:} The global large model $\Theta_i^{(large)}$;\\
      \vspace{-4pt}
      \hrulefill\\
      \COMMENT Train the Local-LiGO for $E_l$ epochs.
      \FOR{$E_l$ epochs}
      \STATE Train Local-LiGO $\mathbf{M}_i^{(local)}$ on local data $\mathcal{D}_i$ with equation~\ref{eq:local_opt_goal} as the goal and equation~\ref{eq:SGD1} as the gradient descent rule;
      \ENDFOR\\
      \vspace{-4pt}
      \hrulefill\\
      \COMMENT{Train and upload the Global-LiGO for $R$ rounds.\\}
      \FOR{$r=1$ to $R$}
      \FOR{$E_g$ epochs}
      \STATE Train $\mathbf{M}_i^{(global)}$ on local data $\mathcal{D}_i$ with equation~\ref{eq:global_opt_goal} as the goal and equation~\ref{eq:SGD2} as the gradient descent rule;
      \ENDFOR
      \STATE Upload its Global-LiGO $\mathbf{M}_i^{(global)}$ to parameter server;
      \STATE $\mathbf{M}_i^{(global)}\leftarrow \text{the aggregated result from parameter server}$ ;
      \ENDFOR
      \STATE Obtain global large model $\Theta_i^{(large)}$ via $\mathbf{M}_i^{(global)}$;
    \end{algorithmic}
    \label{algo:workflowClient}
  \end{algorithm}

  \begin{algorithm}[!ht]
    \caption{Fed-Grow with Dual-LiGO for Parameter Server}
    \begin{algorithmic}[1]
      \FOR{$r=1$ to $R$}
      \STATE Receive $M_i^{(global)}$ from all clients; \\
      \STATE $\mathbf{M}^{(global)} \gets \sum_{i=1}^n \frac{\vert \mathcal{D}_i \vert}{\sum_{j=1}^n \vert \mathcal{D}_j \vert} \mathbf{M}_i^{(global)}$; \\
      \STATE Send $\mathbf{M}^{(global)}$ to all clients;\\
      \ENDFOR
    \end{algorithmic}
    \label{algo:workflowServer}
  \end{algorithm}

  \section{Details of Experimental Setup}\label{appendix:expsetup}

  \subsection{Dataset Descriptions}\label{appendix:datasetintro}
  \begin{itemize}
    \item \textbf{20 Newsgroups}: The 20 Newsgroups dataset~\cite{20news} is a collection of approximately 20,000 newsgroup documents, partitioned nearly across 20 different newsgroups.
    \item \textbf{AG News}: The AG News dataset~\cite{agnews} is a sub-dataset of AG's corpus of news articles, containing 30,000 training and 1,900 test samples per class, with a total of 120,000 training samples and 7,600 testing samples.
    \item \textbf{WikiNER}: The WikiNER dataset~\cite{wikiner} is a large collection of sentences from Wikipedia, predicting 4 tags: 'PER' for person names, 'LOC' for location names, 'ORG' for organization names, and 'MISC' for other names.
    \item \textbf{CIFAR-10 and CIFAR-100}: The CIFAR-10 and CIFAR-100 datasets~\cite{cifar10and100} are labeled subsets of the 80 million tiny images dataset, designed for image classification tasks.
    \item \textbf{Flowers-102}: The Flowers-102 dataset~\cite{flowers102} is an image classification dataset consisting of 102 flower categories commonly occurring in the United Kingdom.
  \end{itemize}

  \subsection{Configurations of Different Transformer Models}\label{appendix:modelconfig}

  The detailed architecture of small models $\Theta^{(small)}$, intermediate models $\Theta^{(inter)}$, and large models $\Theta^{(large)}$ can be found in Table~\ref{tab:models}. For both BERT and ViT, we use the configuration shown in Table~\ref{tab:models}. For each client's small model, we randomly and uniformly sample one of the three heterogeneous model architectures in Table~\ref{tab:models} and train it as a pre-trained small model for that client.

  \begin{table}[ht]
    \centering
    \caption{Configurations of different transformer models.}

    \begin{tabular}{cccc}
      \toprule
      \textbf{Name} & \textbf{\#Hidden} & \textbf{\#Layers} & \textbf{\#Heads} \\
      \midrule
      small\#1      & 256               & 2                 & 8                \\

      small\#2      & 256               & 3                 & 8                \\

      small\#3      & 256               & 4                 & 8                \\

      intermediate  & 320               & 4                 & 8                \\

      large         & 384               & 6                 & 8                \\
      \bottomrule 
    \end{tabular}
    \label{tab:models}
  \end{table}
  \subsection{Implementation Details}\label{appendix:impdetails}
  Our experimental platform is built upon Flower~\cite{flower} and FedNLP~\cite{fednlp} with Pytorch as the training backend. All simulations are conducted on $4\times$ NVIDIA RTX 4090 GPUs with 320GB memory. We conduct our simulations under various conditions to thoroughly evaluate the performance of our models. Our simulations span 6 different tasks, providing a comprehensive evaluation of our models' performance across diverse domains. For different datasets, we set different hyper-parameters $E_l$, $E_g$, and $R$ due to the varying dataset sizes. For each client, we train the small model on the local dataset for $E_l$ rounds as the pre-trained small model. Following~\cite{noniid,fednlp}, we use the Dirichlet distribution with $\beta = 0.5$ to generate non-IID data and $\beta = 100$ to generate IID data, with a fixed random seed. We test our models with different numbers of clients, specifically 10 and 20 clients. The detailed configurations for each task are provided in Table~\ref{tab:hyperparameters}. We referenced the code of \cite{1,2,3,4}.

\subsection{Metrics}
  We use several key metrics to evaluate the performance of our models. For tasks involving image and text classification, we report the \textit{top-1 accuracy} on the test set. For the sequence tagging task, we report \textit{precision} on the test set. In addition to these task-specific metrics, we assess the stability of our models' performance across different clients by calculating and reporting the \textit{standard deviation} of the aforementioned metrics. Moreover, we report the number of \textit{trainable parameters} and \textit{communication cost}, to reflect the resource consumption on computation and communication resources.
  
\begin{table}[ht]
  \centering
  \caption{Hyperparameters for each task.}

  \begin{tabular}{ccccccc}
    \toprule
    \textbf{Task} & \textbf{Batch Size} & \textbf{LR} & $E_l$ & $E_g$ & $R$ & \textbf{Optimizer} \\
    \midrule
    20news        & 32                  & 3e-4        & 100   & 10    & 200 & AdamW              \\

    agnews        & 32                  & 1e-4        & 10    & 1     & 100 & AdamW              \\

    wikiner       & 32                  & 1e-4        & 50    & 3     & 100 & AdamW              \\

    CIFAR-10      & 64                  & 5e-4        & 50    & 4     & 200 & AdamW              \\

    CIFAR-100     & 64                  & 5e-4        & 50    & 4     & 200 & AdamW              \\

    Flowers-102   & 64                  & 5e-4        & 200   & 40    & 100 & AdamW              \\
    \bottomrule
  \end{tabular}

  \label{tab:hyperparameters}
\end{table}

\section{Additional Experimental Results}
\subsection{Curves of Test Accuracy/Precision During Training Process}\label{appendix:learningcurves}

\begin{figure*}[ht]
  \centering
  \begin{subfigure}{0.32\linewidth}
    \includegraphics[width=\linewidth]{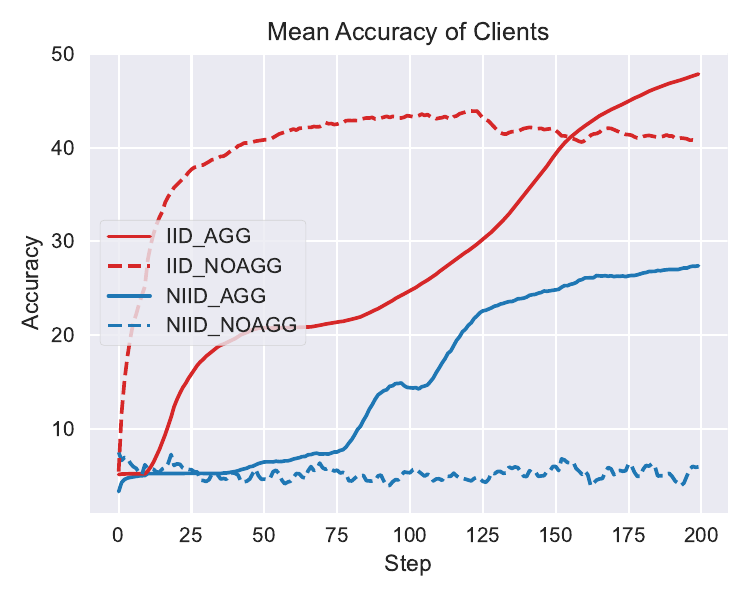}
    \caption{20news}
  \end{subfigure}
  \begin{subfigure}{0.32\linewidth}
    \includegraphics[width=\linewidth]{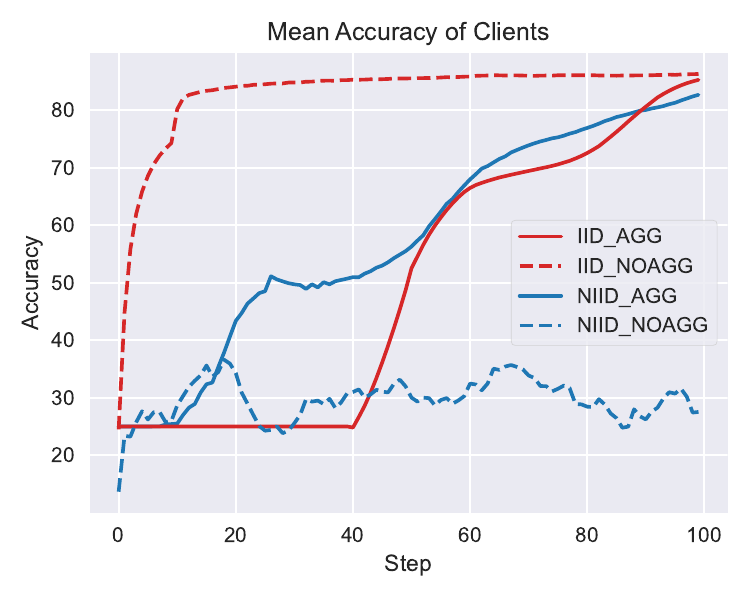}
    \caption{agnews}
  \end{subfigure}
  \begin{subfigure}{0.32\linewidth}
    \includegraphics[width=\linewidth]{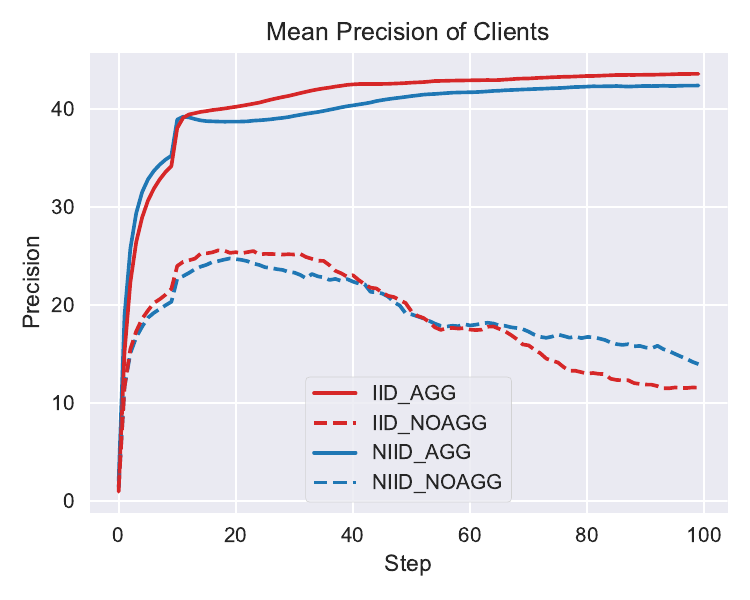}
    \caption{wikiner}
  \end{subfigure}
  \begin{subfigure}{0.32\linewidth}
    \includegraphics[width=\linewidth]{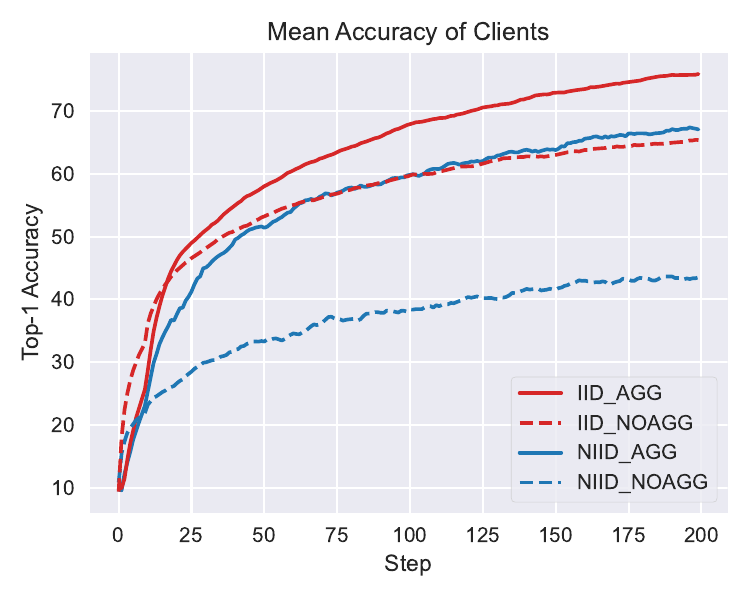}
    \caption{CIFAR-10}
  \end{subfigure}
  \begin{subfigure}{0.32\linewidth}
    \includegraphics[width=\linewidth]{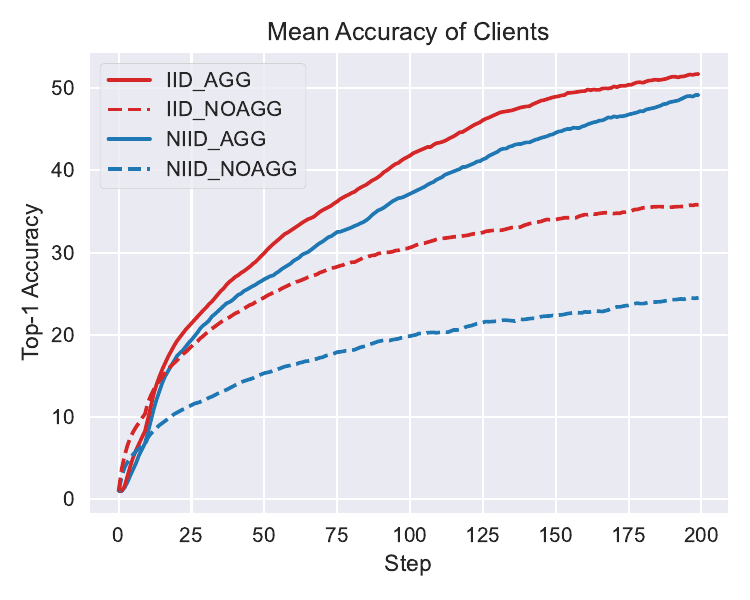}
    \caption{CIFAR-100}
  \end{subfigure}
  \begin{subfigure}{0.32\linewidth}
    \includegraphics[width=\linewidth]{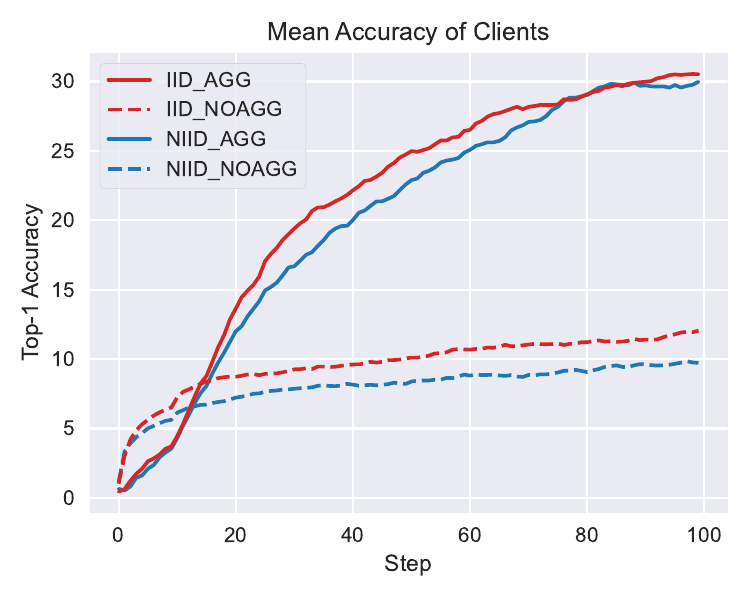}
    \caption{Flowers-102}
  \end{subfigure}

  \caption{Comparison of Dual-LiGO (Agg) and NoAgg on six datasets under IID and non-IID settings with 20 clients.}
  \label{fig:mainResults2}
\end{figure*}

Fig~\ref{fig:mainResults2} exhibit evolution of the mean accuracy/precision over global communication stpes in 20-client settings. From which, we can find that our method outperforms baseline in terms of convergence speed and final performance. The baseline methods are more prone to overfitting (Fig~\ref{fig:mainResults2} (c)) or even failing to converge (Fig~\ref{fig:mainResults2} (b)) due to the reduced amount of data shared by each client, our method effectively avoids these problems.

\begin{figure*}[ht]
  \centering
  \begin{subfigure}{0.32\linewidth}
    \includegraphics[width=\linewidth]{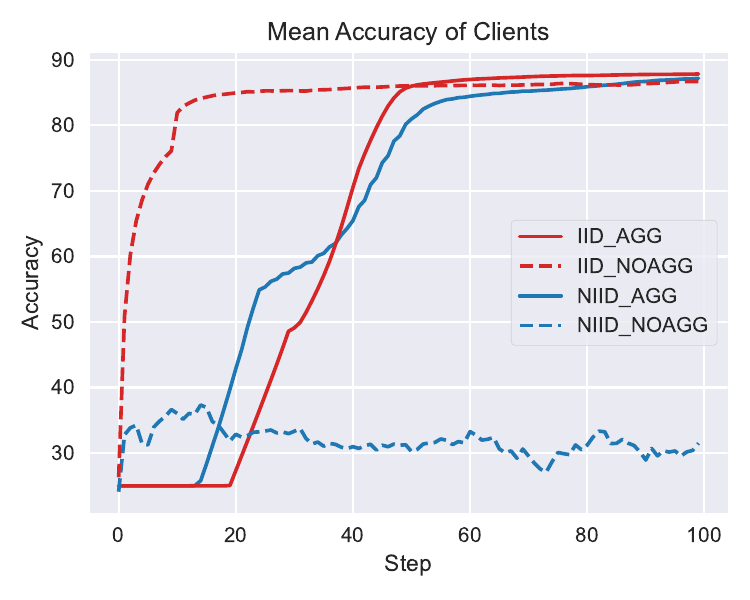}
    \caption{30 clients}
  \end{subfigure}
  \begin{subfigure}{0.32\linewidth}
    \includegraphics[width=\linewidth]{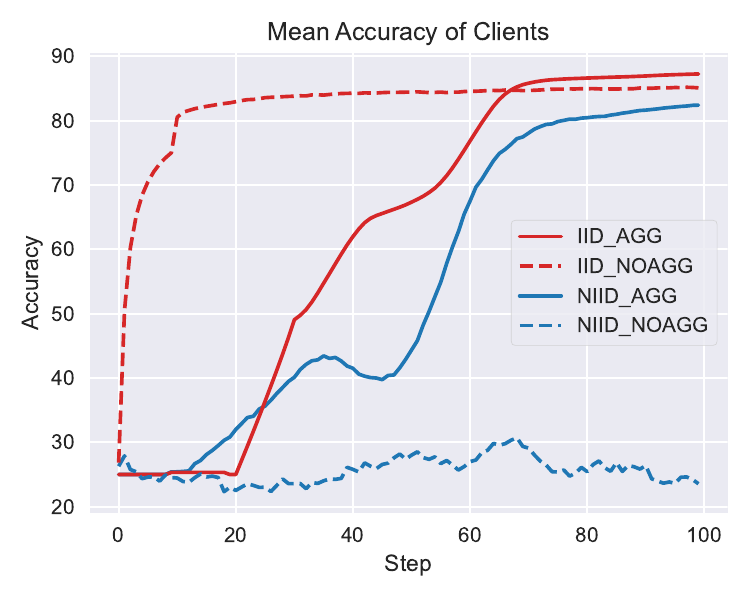}
    \caption{40 clients}
  \end{subfigure}
  \begin{subfigure}{0.32\linewidth}
    \includegraphics[width=\linewidth]{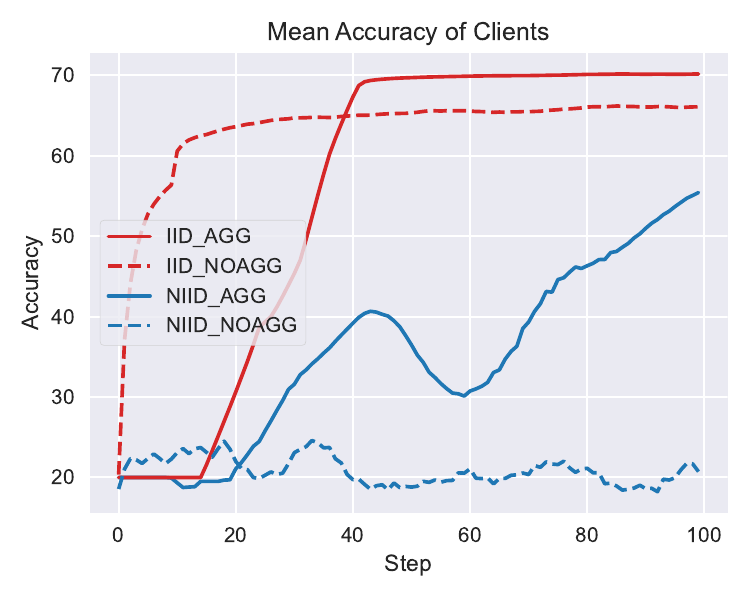}
    \caption{50 clients}
  \end{subfigure}
  \caption{Comparison of Dual-LiGO (Agg) and NoAgg on agnews under IID and non-IID settings with 30-50 clients.}
  \label{fig:differentclientnumagnews}
\end{figure*}

\begin{figure*}[ht]
  \centering
  \begin{subfigure}{0.32\linewidth}
    \includegraphics[width=\linewidth]{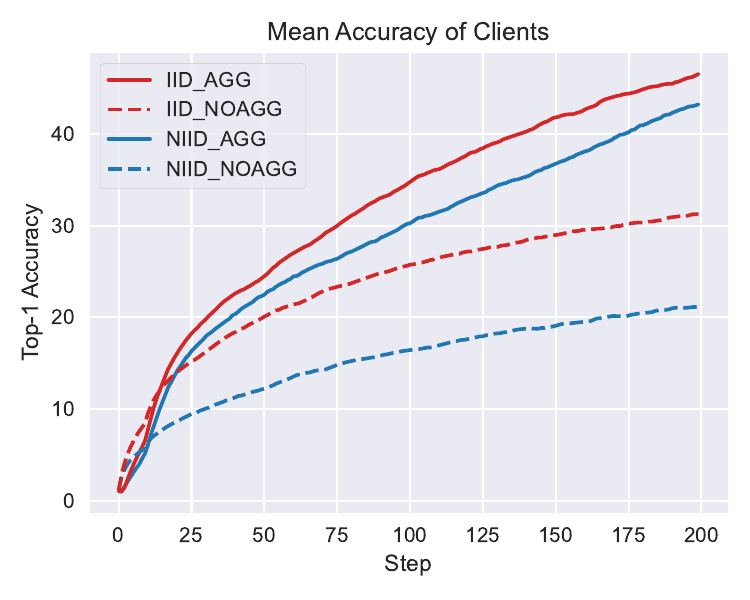}
    \caption{30 clients}
  \end{subfigure}
  \begin{subfigure}{0.32\linewidth}
    \includegraphics[width=\linewidth]{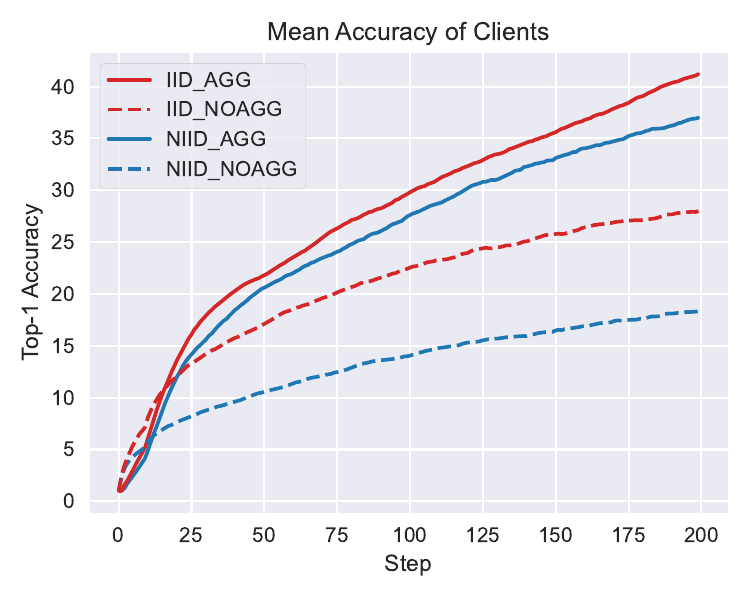}
    \caption{40 clients}
  \end{subfigure}
  \begin{subfigure}{0.32\linewidth}
    \includegraphics[width=\linewidth]{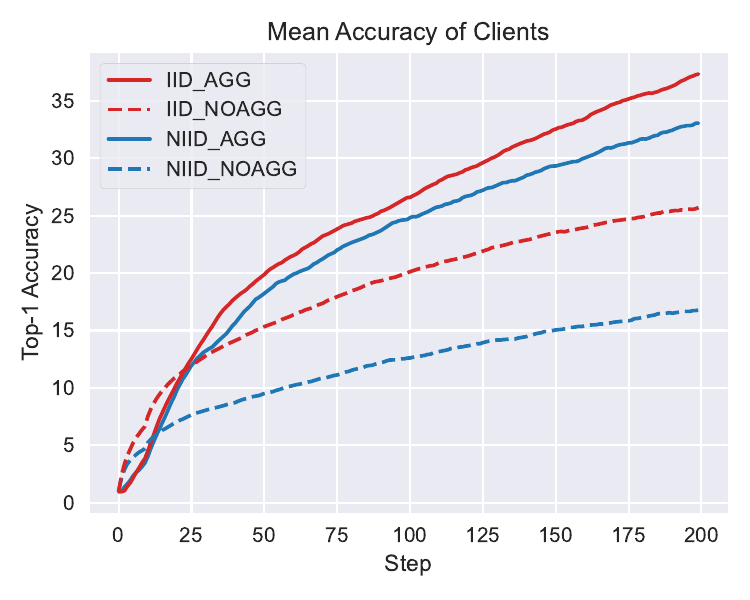}
    \caption{50 clients}
  \end{subfigure}
  \caption{Comparison of Dual-LiGO (Agg) and NoAgg on CIFAR-100 under IID and non-IID settings with 30-50 clients.}
  \label{fig:differentclientnumcifar100}
\end{figure*}

Fig~\ref{fig:differentclientnumagnews} and Fig~\ref{fig:differentclientnumcifar100} show the performance of our method on a larger number of clients (30-50 clients). As the number of clients increases, less data is stored on each client, resulting in slower convergence, but our method still significantly outperforms baseline in terms of convergence speed and final performance.

\subsection{Effects of Small Model Heterogeneity}\label{appendix:effectofhetrogeneity}
In the original experiment, we designed three heterogeneous small models. In order to verify the robustness of our method to the degree of heterogeneity of the small models, in this section, we designed 2 additional heterogeneous cases and ran the experiments in the CIFAR-100 dataset, IID, 10-client settings. The model heterogeneity cases are shown in Table~\ref{tab:hetromodels1}, Table~\ref{tab:hetromodels2} and the final experimental results are presented in Table~\ref{tab:heteroexpresults}. From the table, we can see that regardless of the heterogeneity case, our method has a significant improvement over the baseline method, which proves the tolerance of our method for small model heterogeneity.

\begin{table}[ht]
  \centering
  \caption{Configurations of additional heterogeneity case 1.}

  \begin{tabular}{cccc}
    \toprule
    \textbf{Name} & \textbf{\#Hidden} & \textbf{\#Layers} & \textbf{\#Heads} \\
    \midrule
    small\#1      & 256               & 2                 & 8                \\

    small\#2      & 256               & 3                 & 8                \\

    small\#3      & 256               & 4                 & 8                \\

    small\#4      & 192               & 2                 & 8                \\

    small\#5      & 192               & 3                 & 8                \\

    small\#6      & 192               & 4                 & 8                \\

    intermediate  & 320               & 4                 & 8                \\

    large         & 384               & 6                 & 8                \\
    \bottomrule
  \end{tabular}
  \label{tab:hetromodels1}
\end{table}

\begin{table}[!ht]
  \centering
  \caption{Configurations of additional heterogeneity case 2.}

  \begin{tabular}{cccc}
    \toprule
    \textbf{Name} & \textbf{\#Hidden} & \textbf{\#Layers} & \textbf{\#Heads} \\
    \midrule
    small\#1      & 256               & 2                 & 8                \\

    small\#2      & 256               & 3                 & 8                \\

    small\#3      & 256               & 4                 & 8                \\

    small\#4      & 256               & 5                 & 8                \\

    small\#5      & 256               & 6                 & 8                \\

    intermediate  & 320               & 7                 & 8                \\

    large         & 384               & 8                 & 8                \\
    \bottomrule
  \end{tabular}
  \label{tab:hetromodels2}
\end{table}

\begin{table}[!ht]
  \centering
  \caption{Mean accuracy of Dual-LiGO (Agg) and NoAgg on the CIFAR-100 dataset under IID, 10-client settings with different heterogeneity of small models.}
  \begin{tabular}{cccc}
    \toprule
    \multirow{3}{*}{\textbf{Heterogeneity Case}} & \multicolumn{3}{c}{\textbf{Accuracy (\%)} $\uparrow$}                                     \\ \cmidrule{2-4}
                                                 & Agg                                                   & NoAgg  & $\Delta$                 \\ \midrule
    Original Case                                & 53.178                                                & 38.127 & \textcolor{red}{15.051}  \\
    Additional Case 1                            & 57.812                                                & 38.306 & \textcolor{red}{19.506 } \\
    Additional Case 2                            & 53.949                                                & 47.568 & \textcolor{red}{6.381}   \\ \bottomrule
  \end{tabular}
  \label{tab:heteroexpresults}
\end{table}

\subsection{Comparison with FedAvg with Only Large Model}
In this section, we compare our approach with FedAvg, specifically, we resize the large model to ensure roughly the same amount of communication as our approach, and then train the large model, upload and aggregate it (There are no small models and intermediate models stored in clients); Table~\ref{tab:cmpwithfedavg} shows the results of the comparison in 20news dataset, IID and 10-client settings. Due to the limitation of communication resources, the parameter number of the large model is limited, and the performance of our algorithm has a significant advantage over directly training a large model under the same communication resources. Moreover, compared to directly training a large model, our approach is able to utilize the implicit knowledge of the pre-trained small model on each client to further accelerate the training process.

\begin{table}[!ht]
  \caption{Comparative results of communication cost and accuracy between FedAvg and our approach on the 20news dataset under IID conditions with 10 clients.}
  \centering
  \begin{tabular}{ccc}
    \toprule
    \textbf{Method} & \textbf{Comm Cost} & \textbf{Accuracy (\%) $\uparrow$} \\ \midrule
    FedAvg          & 4.034M             & 53.638                            \\
    Ours            & 4.178M             & \textbf{65.295}                   \\ \bottomrule
  \end{tabular}
  \label{tab:cmpwithfedavg}
\end{table}

\subsection{Visualization of Intermediate Models' Similarity}
\begin{figure*}[ht]
  \centering
    \includegraphics[width=\linewidth]{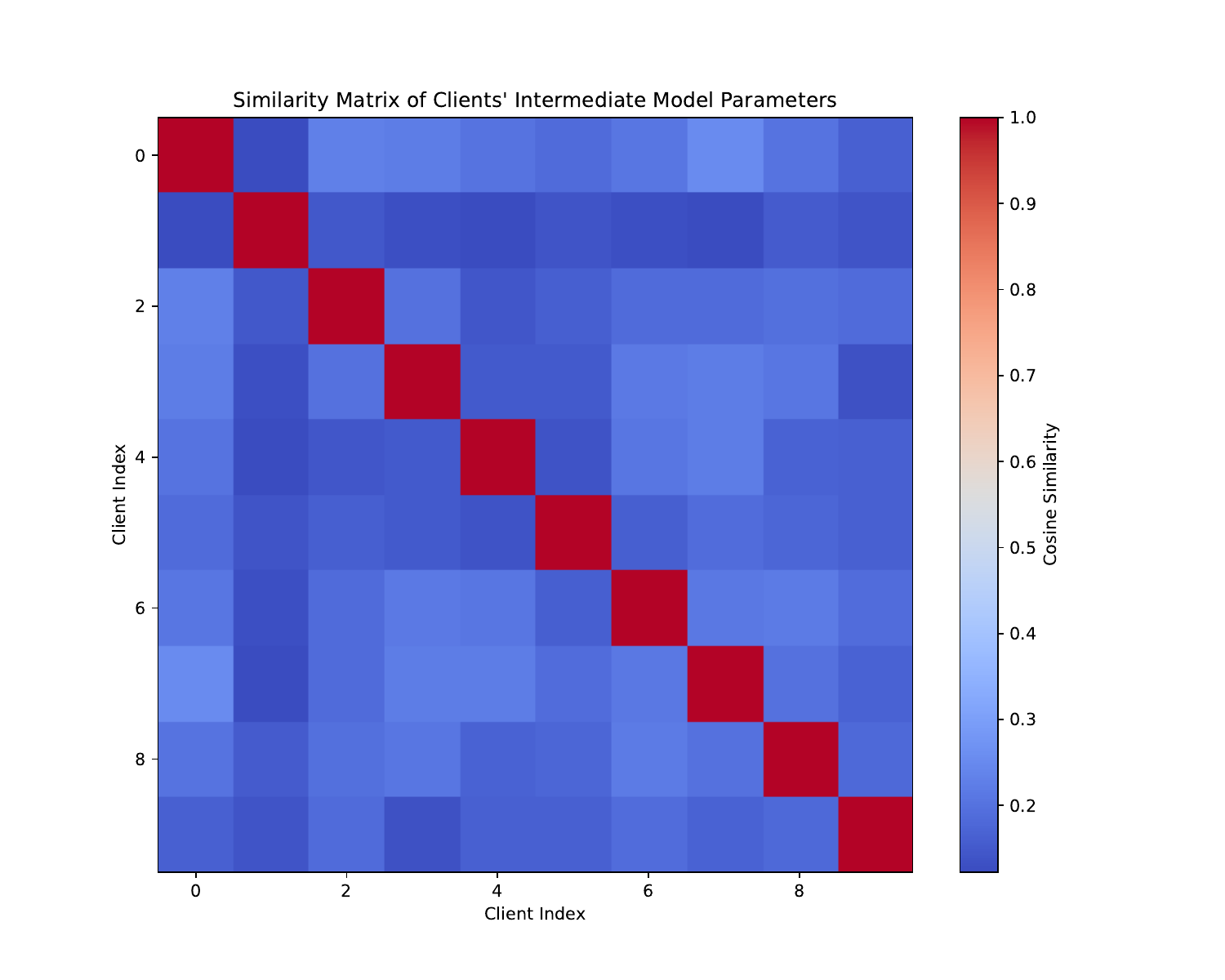}
  \caption{The cosine similarity between intermediate models in different clients in agnews dataset, non-IID and 10-clients setting.}
  \label{fig:cossim}
\end{figure*}

In our investigation of privacy preservation within our proposed framework, we explore the pair-wise similarity between intermediate models contributed by different clients in agnews dataset, non-IID and 10-clients setting. The similarity matrix, computed using cosine similarity is shown in Figure~\ref{fig:cossim}.

According to~\cite{privacy}, clients with highly similar models (values close to 1) may inadvertently leak information about their individual data distributions. Such similarity could lead to privacy risks, including membership inference attacks. On the other hand, clients with diverse models (values closer to 0) enhance privacy by introducing aggregated noise and making it harder for adversaries to pinpoint specific contributions. In our case, the similarity between the clients' intermediate models is very low, which indicates the advantage of our approach for privacy preservation. In addition to this, we still need to emphasize that during the transmission stage of our models, only Global-LiGO is transmitted, and it is very difficult for an attacker to access the parameters of our intermediate or small models as well as the training data of each client through traditional attack measures.

\section{Limitations}\label{appendix:limitations}
Despite the innovative approach and promising results of Fed-Grow with Dual-LiGO, we acknowledge certain limitations that warrant further investigation.

In the spirit of continuous improvement, we note that the current iteration of Fed-Grow shows room for enhancement in scenarios involving 20-client agnews datasets with non-IID settings (can be found in Table~\ref{tab:mainResults}). This observation opens up opportunities to deepen our understanding of data diversity's impact on model performance. Our forthcoming research will be dedicated to exploring these dynamics further, with the aim of fortifying the robustness of our framework across varied client environments.

The journey of a global large model to its full potential includes a pivotal step of client-specific fine-tuning. This process is key to unlocking a model's robust performance tailored to individual datasets. Recognizing that computational resources are a valuable commodity, we are motivated to explore more efficient fine-tuning methodologies. Our future work is poised to focus on this endeavor, aspiring to make the fine-tuning process as seamless as possible for all clients, regardless of their resource availability.

\clearpage
\end{document}